\def\year{2020}\relax
\documentclass[letterpaper]{article} 
\usepackage{aaai20}  
\usepackage{times}  
\usepackage{helvet} 
\usepackage{courier}  
\usepackage[hyphens]{url}  
\usepackage{graphicx} 
\usepackage{graphicx}  
\usepackage{latexsym,amssymb,gensymb,amstext,amsfonts,amsmath,mathtools,bm}
\usepackage{lipsum}
\usepackage{booktabs}
\usepackage{comment}
\usepackage{array}

\newcommand{\sssection}[1]{\noindent\textbf{#1}\quad}
\newcommand{\mathfs}[0]{\fontsize{8.5}{9.5}\selectfont}
\newcommand{\mathfsapp}[0]{\fontsize{7.0}{8.0}\selectfont}
\newcommand{\tsnefs}{0.56}

\title{Invariant Representations through Adversarial Forgetting}

\author{
Ayush Jaiswal, Daniel Moyer, Greg Ver Steeg, Wael AbdAlmageed, Premkumar Natarajan\\
Information Sciences Institute, University of Southern California\\
  \texttt{\{ajaiswal, gregv, wamageed, pnataraj\}@isi.edu, moyerd@usc.edu}
}

\begin{document}

\maketitle

\begin{abstract}
We propose a novel approach to achieving invariance for deep neural networks in the form of inducing amnesia to unwanted factors of data through a new adversarial forgetting mechanism. We show that the forgetting mechanism serves as an information-bottleneck, which is manipulated by the adversarial training to learn invariance to unwanted factors. Empirical results show that the proposed framework achieves state-of-the-art performance at learning invariance in both nuisance and bias settings on a diverse collection of datasets and tasks.
\end{abstract}

\section{Introduction}
\label{sec:introduction}

Supervised machine learning models learn to associate a target $y$ with underlying factors of data $x$ that are informative of $y$. However, trained models often learn to associate irrelevant factors of $x$ with $y$~\cite{bib:domingos}, e.g., stroke-width of text in optical character recognition. Learning incorrect associations between $y$ and \emph{nuisance} factors leads to overfitting. Models might also learn associations between the target $y$ and factors that are correlated with $y$ in collected training data due to external (sometimes historical) reasons, e.g., age, gender, and race in socio-economical prediction tasks. These \emph{biasing} factors, which are sometimes known \textit{a priori}, make trained models unfair to under-represented groups, posing ethical and legal challenges to the usage of such models. It is, therefore, imperative to develop models that are invariant to both nuisance and biasing factors.

Popular approaches for achieving invariance to undesired factors $s$ have involved training with data augmentation through small perturbations to real $x$ with respect to $s$~\cite{bib:bengio2013}. While these methods have been utilized for training deep neural networks (DNNs)~\cite{bib:data_aug_1,bib:data_aug_2}, in recent years methods have emerged that remove $s$ from the latent representations $\tilde{z}$, penalizing the model for the presence of $s$~\cite{bib:uai,bib:unifai,bib:nnmmd,bib:hcv,bib:vfae,bib:cvib,bib:cai,bib:lfr}. These methods perform better than data augmentation, due to their approach of invariance by \emph{exclusion} rather than \emph{inclusion}~\cite{bib:uai,bib:ropad,bib:niesr}.

We propose a novel framework for invariance in DNNs that promotes removal of information about $s$ from $\tilde{z}$ through an adversarial forgetting mechanism. The working principle of the model is to map data $x$ to an embedding $z$ that encodes everything about $x$ and use a forget-mask to filter out $s$ information from $z$ while retaining information about $y$ to produce the invariant $\tilde{z}$. Specifically, an encoder network generates a latent code $z$ from $x$, which is used to reconstruct $x$ through a decoder. At the same time, a forget-gate network generates a mask $m$ from $x$, which is multiplied elementwise with $z$ to produce $\tilde{z}$. The encoding $\tilde{z}$ is used by a predictor to infer the target $y$. These components of the framework are trained adversarially with a discriminator that aims to predict $s$ from $\tilde{z}$. However, during training, gradients from the discriminator are allowed to only affect the training of the forget-gate. Finally, the framework is augmented with a regularizer that pushes the components $m_i$ of the forget-mask to be close to either $0$ or $1$, inducing disentanglement within the components of $z$ and effective masking of $s$ to produce an invariant representation $\tilde{z}$.

We show that the forgetting mechanism is equivalent to a bound on the mutual information $I(\tilde{z}:z)$ and that coupled with the $y$-prediction task, can be interpreted as an information bottleneck. Further, by the data processing inequality, the forgetting mechanism bounds $I(\tilde{z}:s)$. The generated mask can be manipulated through adversarial training to remove information about $s$ from $z$. Empirical results show that the proposed framework exhibits state-of-the-art performance at inducing invariance to both nuisance and biasing factors across a diverse collection of datasets and tasks.

\section{Related Work}
\label{sec:related_work}

Recent work~\cite{bib:achille2018,bib:vib,bib:cvib} has modeled invariance in supervised DNNs through information bottleneck~\cite{bib:ib}, wherein representations minimize the mutual information $I(x:\tilde{z})$ while maximizing $I(\tilde{z}:y)$. For nuisance variables ($s \perp y$), these methods bring about compression in the latent space, which removes information about $s$ and indirectly minimizes $I(\tilde{z}:s)$. Under optimality, the information bottleneck objective will remove all such nuisance $s$ completely from $\tilde{z}$~\cite{bib:achille2018}. The bottleneck objective is, however, difficult to optimize and has, hence, been approximated using variational inference in prior work~\cite{bib:vib} or implemented as Information Dropout~\cite{bib:infodropout}.

The bottleneck objective cannot remove $s$ from $\tilde{z}$ that are correlated with $y$ (biasing factors). In such cases, it is necessary to force the exclusion of these biasing $s$ from $\tilde{z}$. Methods that employ such mechanisms additionally benefit the removal of nuisance $s$ for which annotations are available. Moyer \textit{et al.}~(\citeyear{bib:cvib}) achieve this by augmenting the bottleneck objective with $I(\tilde{z}:s)$ and optimizing its variational bound, which we call the Conditional Variational Information Bottleneck (CVIB). The Variational Fair Autoencoder (VFAE)~\cite{bib:vfae} optimizes this objective indirectly as a Variational Autoencoder (VAE)~\cite{bib:vae} and uses Maximum Mean Discrepancy (MMD)~\cite{bib:mmd} to boost the removal of undesired $s$ from the latent embedding. The Hilbert-Schmidt Independence Criterion (HSIC)~\cite{bib:hsic} constrained VAE (HCV)~\cite{bib:hcv} uses HSIC to enforce independence between the latent embedding and $s$. NN+MMD~\cite{bib:nnmmd} directly minimizes MMD as a regularizer for DNNs.

Invariance to nuisance variables has also been modeled as an implicit feature selection method within DNNs in the Unsupervised Adversarial Invariance framework (UAI)~\cite{bib:uai,bib:unifai} that splits data representation into an invariant embedding, which is used for predicting $y$, and a nuisance embedding through competitive training between prediction and reconstruction tasks combined with disentanglement between the two embeddings.

The Domain-Adversarial Neural Network (DANN)~\cite{bib:dann} and the Controllable Adversarial Invariance (CAI) model~\cite{bib:cai} indirectly optimize an invariance objective using the gradient-reversal trick~\cite{bib:dann} that penalizes a DNN if it encodes $s$. In the case of invariant representation learning for a supervised prediction task (which is the subject of this work), CAI, DANN and Fader Networks~\cite{bib:fader} reduce to the same model. In contrast, we propose a new forgetting mechanism that more explicitly ``masks'' $s$ out of $z$, which encodes everything about $x$, to generate a new representation $\tilde{z}$ that is maximally informative of $y$ but invariant to $s$. Inspired by (1) the discovery of richer features in DNN classifiers upon augmentation with reconstruction objectives~\cite{bib:recon_2,bib:recon_1}, and (2) the forgetting operation in Long Short-Term Memory (LSTM)~\cite{bib:lstm} cells, the proposed framework adopts the idea of ``discovery and separation of information'' for invariance, which is \emph{foundationally different} from the direct removal of undesired factors as in DANN and CAI.

\section{Invariance through Adversarial Forgetting}
\label{sec:method}

Invariant representation learning aims to produce a mapping of data ($x$) to code ($\tilde{z}$) that is minimally informative of undesired factors of data ($s$) but maximally discriminative for the prediction task ($y$). There are two cases that arise from this formulation -- (1) $s$ is nuisance, i.e., there is little or no information shared between $s$ and $y$ asymptotically (e.g., pose in face recognition), and (2) $s$ contains biasing factors, i.e., there is correlation between $s$ and $y$, but for outside reasons, it is necessary to exclude these biases from the prediction process (e.g., gender, race, etc. in socio-economical prediction tasks using historical data). Invariance to $s$ leads to robust models that generalize better on test data for case (1)~\cite{bib:uai}, and produces fair models that do not incorporate $s$ while making predictions for case (2).

\begin{figure*}[t]
\centering
\includegraphics[width=0.75\linewidth]{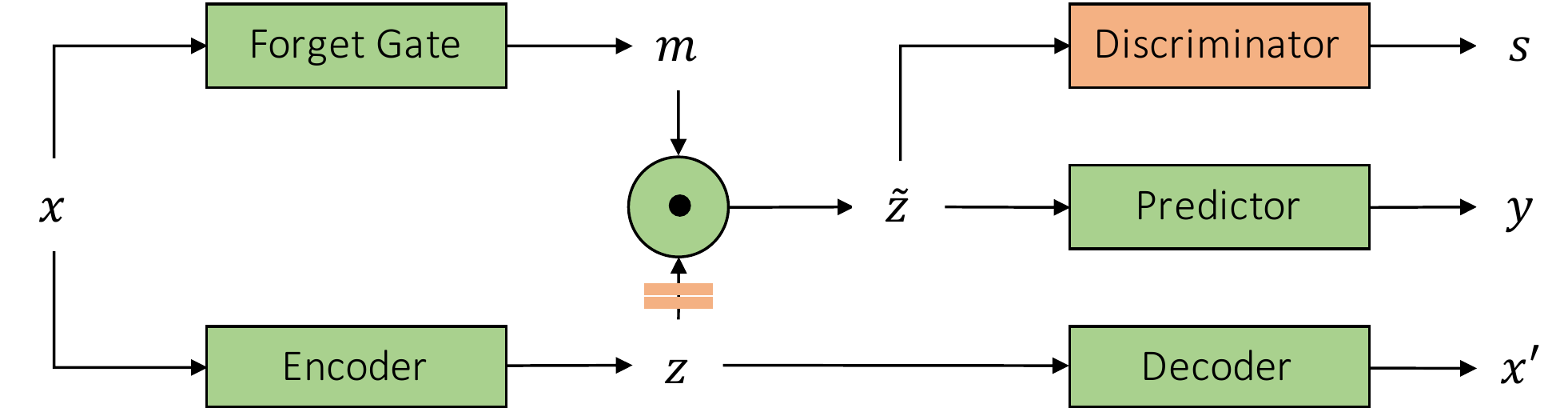}
\caption{\label{fig:framework}Adversarial forgetting framework for invariant representation learning}
\end{figure*}

We induce invariance to $s$ within a DNN using a novel approach of adversarial forgetting, which is inspired by forget-gates in Long Short-Term Memory (LSTM) cells in recurrent neural networks~\cite{bib:lstm}. Within the proposed framework, the model learns to embed everything about a data sample into an intermediate representation, which is transformed into an invariant representation through multiplication with a mask generated by an adversarial forgetting mechanism. Figure~\ref{fig:framework} shows the complete framework design. Data samples $x$ are encoded into an intermediate representation $z$ using an encoder $E$, while a forget-mask $m$ ($m_i \in (0, 1)$) is simultaneously produced from $x$ through a forget-gate network $F$. The invariant representation $\tilde{z}$ is then computed as the element-wise multiplication  $\tilde{z} = z \odot m$. This is similar to how forget gates are used in LSTMs to ``forget'' certain information learned from past data. A decoder $R$ is used to reconstruct $x$ from $z$, such that $E$ learns to encode everything about $x$ into $z$. A predictor $P$ infers $y$ from $\tilde{z}$, while an adversarial discriminator $D$ tries to predict $s$ from $\tilde{z}$. Hence, the combined objectives of $P$ and $D$ aim to allow only factors of $x$ that are predictive of $y$ but not of $s$ to pass from $z$ to $\tilde{z}$.

The complete framework is trained with an adversarial objective such that the discriminator is pitted against all the other modules, as depicted with colors in Figure~\ref{fig:framework}. The discriminator is allowed a more active role in the development of forget-masks $m$ by allowing adversarial gradients from the discriminator to only flow to the forget-gate network and not to the encoder during training. This is illustrated in the figure with the break on the arrow between $z$ and the multiplication operation. In order to further encourage the development of masks that truly filter out some information from $z$ but retain everything else, a mask regularizer, in the form of $m^{\text{T}}(1-m)$, is added to push components of $m$ to either $0$ or $1$. We found the mask regularizer to always improve results in our experiments. The complete training objective can be written as shown in Equation~\ref{eq:objective}.
{
\mathfs
\begin{align}
\min_{E, F, P, R} \ \ \max_{D} \ &J(E, F, P, R, D); \ \ \text{where:} \nonumber \\
J(E, F, P, R, D) = \ &L_{y} \bigl(y, P(\tilde{z}) \bigr) + \rho L_{x} \bigl(x, R(z)\bigr) \nonumber \\
&~~~~+ \ \delta L_{s} \bigl(s, D(\tilde{z}) \bigr) + \lambda m^{\text{T}}(1-m)
\label{eq:objective}
\end{align}
}
The losses $L_{y}$ and $L_{s}$ are implemented as cross-entropy while mean squared error is used for $L_{x}$ in our experiments. The proposed model is trained using a scheduled update scheme similar to the training mechanism of the UAI model~\cite{bib:unifai}. Hence, the weights of the discriminator are frozen when updating the rest of the model and vice versa. The adversarial training of the proposed model would benefit from training the discriminator to convergence before any update to the other modules~\cite{bib:unifai}. However, in practice, this is infeasible and training the discriminator much more frequently than the rest of the model (depending on the nature of the prediction task and the dataset) is sufficient to achieve good performance. This is especially true because the training of the discriminator is resumed from its previous state rather than starting from scratch after every update to the other modules. Therefore, the weights of the discriminator and the rest of the model are updated in the frequency ratio of $k:1$. We found $k=10$ to work well in our experiments.

The model training does not incorporate the popular approach of gradient-reversal and instead follows~\cite{bib:unifai}. The targets of the discriminator $D$ are set to the ground-truth $s$ labels while updating $D$, but to random $s$ values (sampled from the empirical $s$-distribution) when the parameters of the rest of the model are updated. Hence, $D$ tries to predict the correct $s$ during its training phase, but the rest of the model is updated to elicit random-chance performance at $s$-prediction, leading to the desired invariance to $s$. The model was implemented in Keras with TensorFlow backend. The Adam optimizer was used with $10^{-4}$ learning rate and $10^{-4}$ decay. The hyperparameters $\rho$, $\lambda$, and $\delta$ were tuned through grid search in powers of $10$.

The proposed framework can also be extended to multi-task settings, with $z$ treated as the common encoding for tasks involving prediction of targets $\{y^{(1)}, y^{(2)},\ldots,y^{(n)}\}$ with corresponding undesired factors $\{s^{(1)}, s^{(2)},\ldots,s^{(n)}\}$. Forget-gates $F^{(j)}$ are added to the framework, one for each prediction task $y^{(j)}$, to generate associated masks $m^{(j)}$ and invariant representations $\tilde{z}^{(j)}$ from $z$ through adversarial training with discriminators $D^{(j)}$, each of which tries to predict $s^{(j)}$ from $\tilde{z}^{(j)}$. Thus, the multi-task extension of the proposed model is intuitive and straightforward.

\section{Characterizing Forgetting with Forget-gate}
\label{sec:analysis}

Forget-gates were introduced as components of LSTMs, where they cause ``forgetting'' of information from the past in the recurrence formulation conditioned on the input at a given step as well as the existing state. A forget-gate typically produces a mask $m$ with $m_i \in (0, 1)$ that is multiplied elementwise with a latent encoding within an LSTM cell. Thus, the forget-gate can scale or remove information in each dimension of the encoding \emph{but not add information to it}. Inspired by this formulation, we employ forget-gates to induce invariance to $s$, i.e., to ``forget'' $s$-related information. In this section, we characterize the erasure properties of the forget-gate in the proposed framework. Intuitively, if a mask element $m_i = 0$, the information passed from $z_i$ to $\tilde{z}_i$ is also zero; likewise, if $m_i = 1$, the information passed is complete. We characterize here the behavior for $m_i \in (0, 1)$, showing that under reasonable assumptions, there is a non-trivial forget regime besides zero. We show that the forget-gate \textit{acts as} an information bottleneck, which can be manipulated for invariance to specific $s$.

\begin{figure*}
\centering
\includegraphics[width=\tsnefs\textwidth]{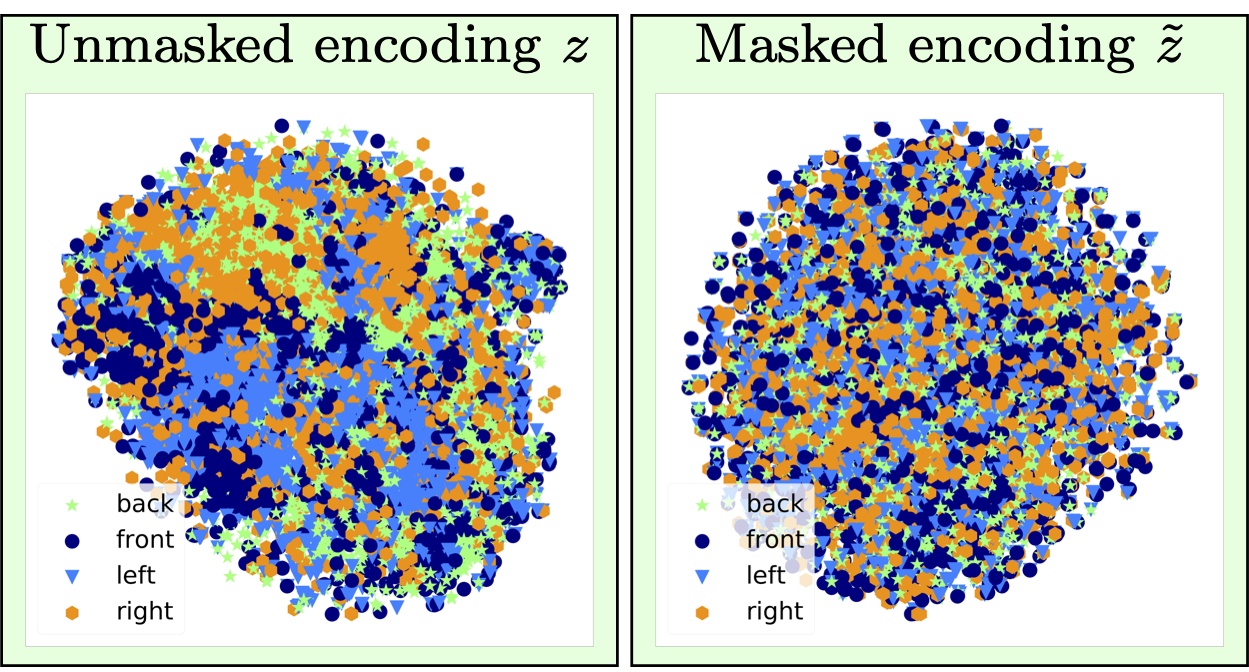}
\caption{Chairs Dataset -- t-SNE visualization of $z$ and $\tilde{z}$ labeled with orientation class ($s$). Visualization with chair-type ($y$) annotations are not shown because there are 1,393 $y$ classes. The invariant encoding $\tilde{z}$ shows no clustering by orientation as $s$ is masked out of $z$, which exhibits $s$-grouping.}
\label{fig:tsne_chairs}
\end{figure*}

\subsection{Forget-gate}
\label{subsec:forget_bottleneck}

The proposed model generates a $d$-dimensional encoding $z$ of $x$ and a forgetting mask $m$ with components $m_i \in (0, 1)$. These are multiplied element-wise to produce $\tilde{z} = z \odot m$. We consider the multiplication as a noisy operation $\tilde{z} = z\odot m + \varepsilon$ with a small $\varepsilon$ in order to facilitate a theoretical analysis of ``forgetting''. We discuss the practicalities of $\varepsilon$ later in Section~\ref{subsec:epsilon_practical}. Assuming $\varepsilon \sim \mathcal{N}(0,\sigma_\varepsilon I)$, we get $P(\tilde{z}|z) \sim \mathcal{N}(z\odot m,\sigma_\varepsilon I)$. In each dimension, we get:
{
\mathfs
\begin{align}
I(\tilde{z}_i : z_i) &= H(\tilde{z}_i) - H(\tilde{z}_i | z_i) = H(\tilde{z}_i) - H(\varepsilon_i) \nonumber \\
&= H(\tilde{z}_i) - \frac{1}{2}  \log(\text{Var}(\varepsilon_i)) - \frac{1}{2}  \log (2\pi e) 
\end{align}
}
where $H(\varepsilon_i)$ is constant with respect to $z_i$ and $m_i$. Thus, the information passed from $z_i$ to $\tilde{z}_i$ is proportional to $H(\tilde{z}_i)$. Assuming that $\text{Var}(z_i)$ is defined, the max-entropy Gaussian upper bound gives us the following in each dimension of the embedding~\cite{bib:c-and-t-IT}:
{
\mathfs
\begin{align}
    H(\tilde{z}_i) &= H(m_i z_i + \varepsilon_i) \\
    &\leq \frac{1}{2} \log(\text{Var}(m_i z_i) + \text{Var}(\varepsilon_i))  + \frac{1}{2}\log (2\pi e)
\end{align}
}
If $m$ is non-random, the mutual information is:
{
\mathfs
\begin{align}
    I(\tilde{z}_i : z_i) &= H(\tilde{z}_i) - H(\varepsilon_i) \\
    &\leq \frac{1}{2} \log( m_i^2\text{Var}(z_i) + \text{Var}(\varepsilon_i) )  - \frac{1}{2} \log( \text{Var}(\varepsilon_i) ) \label{eq:bound}
\end{align}
}
When $m_i \rightarrow 1$, assuming that $\text{Var}(\varepsilon_i) \ll \text{Var}(z_i)$, $H(\tilde{z}_i) \approx H(z_i)$. As $m_i \rightarrow 0$, $\tilde{z}_i = \varepsilon_i$, $H(\tilde{z}_i) \rightarrow H(\varepsilon_i)$ and $I(\tilde{z}_i : z_i) \rightarrow 0$. Importantly, when $m_i < \sqrt{Var(\varepsilon_i)}$ yet away from zero, the information loss is still non-trivial. The result in Equation~\ref{eq:bound} was derived assuming that $m$ is fixed. We can extend this bound to random $m$, including those dependent on $x$ as described in Section~\ref{sec:method}, by using the following identity (shown in Appendix~\ref{sec:app:var-id}):
{
\mathfs
\begin{align}
    \text{Var}(m_i z_i) & ~~\leq~~ 2\text{Var}((m_i - \mathbb{E}[m_i])z_i) + 2 \mathbb{E}[m_i]^2 \text{Var}(z_i)
\end{align}
}
This makes the bound on $I(\tilde{z}_i : z_i)$:
{
\mathfs
\begin{align}
I(\tilde{z}_i : z_i) ~~&\leq~~ \frac{1}{2} \log( 2\text{Var}((m_i - \mathbb{E}[m_i])z_i)  + \text{Var}(\varepsilon_i) \nonumber \\ 
&~~~~~~~~~~~~+ 2 \mathbb{E}[m_i]^2 \text{Var}(z_i)) - \frac{1}{2} \log( \text{Var}(\varepsilon_i) )
\end{align}
}
Though more difficult to interpret, this has approximately the same characteristic as the fixed $m_i$ case. In order to extend this to the multivariate case (from a bound on  $I(\tilde{z}_i : z_i)$ to that on $I(\tilde{z} : z)$), we first note that the max-entropy bound still holds for multivariate Gaussians, and by Hadamard's inequality, we can bound that distribution by its diagonal as:
{
\mathfs
\begin{align}
H(\tilde{z}) ~~&\leq~~ \log \det (\Sigma_{\tilde{z}} + \sigma_\varepsilon I) + \frac{d}{2}\log(2\pi e) \\
~~&\leq~~ \log \det(\Sigma_{\tilde{z}}^{diag} + \sigma_\varepsilon I)  + \frac{d}{2}\log(2\pi e)
\end{align}
}
where $\Sigma_z^{diag}$ has only the diagonal elements of $\Sigma_z$ and zero elsewhere. This gives us the bound:
{
\mathfs
\begin{align}
I(\tilde{z} : z) ~~\leq~~ \sum\mathop{}_{\mkern-5mu i} \log(\text{Var}(m_i z_i) + \text{Var}(\varepsilon_i))
\end{align}
}
The max-entropy bound fails if there are degenerate elements of $z$, i.e. completely duplicate channels, but still holds on subsets of channels without duplicates. We have somewhat abused notation in this section; really, our bound is on $I(\tilde{z} : (z,m))$. While this is less intuitive, the distinction is necessary for data processing inequalities showing, e.g., that $I(\tilde{z} : s)$ is controlled by the forget gate.

\begin{table}
\centering
\caption{\label{tab:chairs}Chairs results (random chance of $s = 0.25$).}
\begin{tabular}{lcc}
\toprule
\textbf{Model} & $A_y$ & $A_s$ \\
\cmidrule(r){1-1} \cmidrule(lr){2-2} \cmidrule(l){3-3}
NN+MMD & 0.73 $\pm$ 0.02 & 0.46 $\pm$ 0.04 \\
VFAE & 0.72 $\pm$ 0.04 & 0.37 $\pm$ 0.02 \\
CAI & 0.68 & 0.69 \\
CVIB & 0.67 $\pm$ 0.01 & 0.52 $\pm$ 0.01 \\
UAI & 0.74 & 0.34 \\
\cmidrule(r){1-1}
\textbf{Ours} & \textbf{0.84 $\pm$ 0.01} & \textbf{0.25 $\pm$ 0.00} \\
$\Delta$ over UAI & 38.5\% & 100\% \\
\bottomrule
\end{tabular}
\end{table}

\begin{figure*}
\centering
\includegraphics[width=\tsnefs\textwidth]{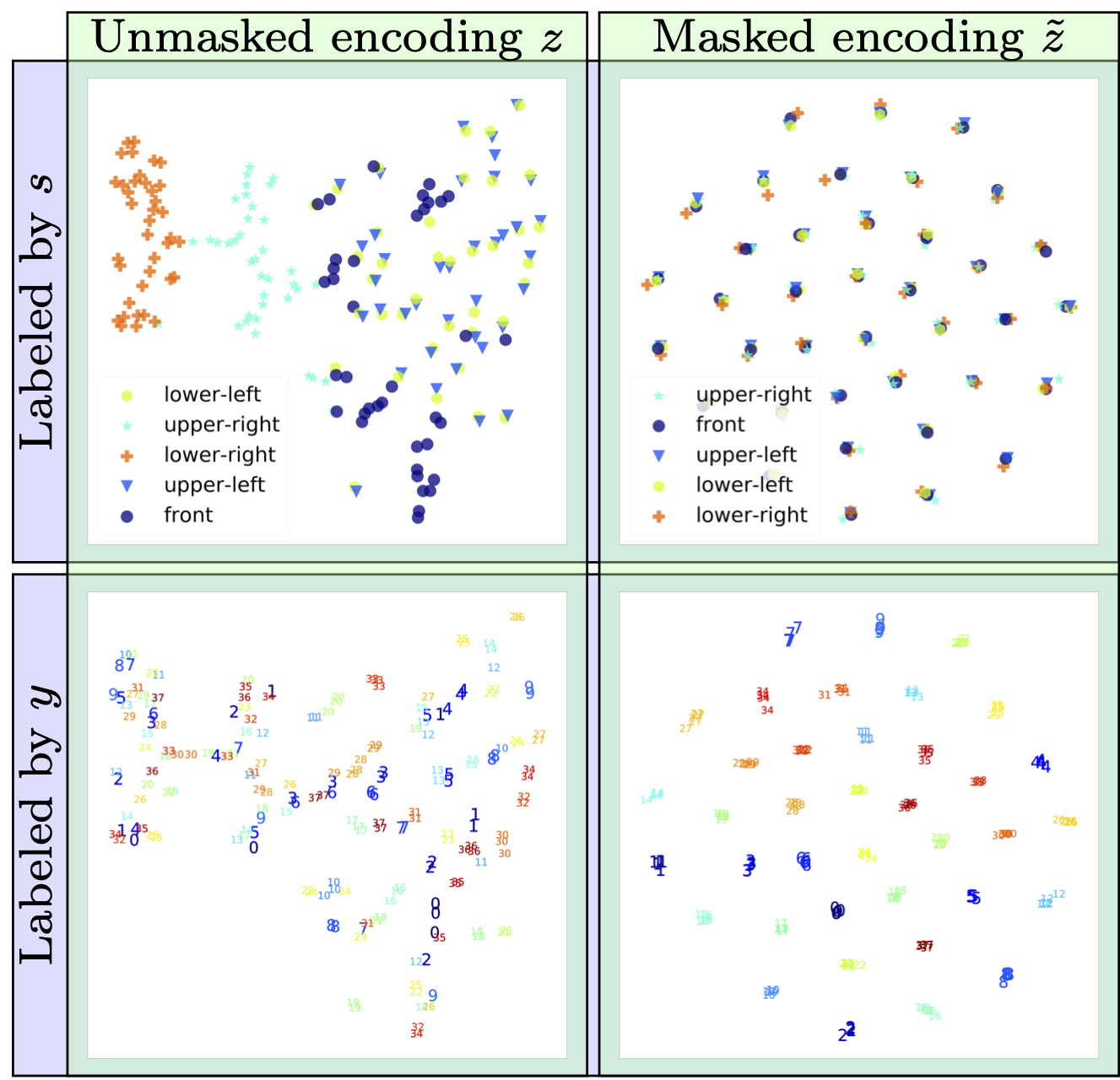}
\caption{Extended Yale-B -- t-SNE visualizations labeled with lighting direction ($s$) and subject-identity ($y$). Subject-identities are marked with numbers and not colors. The invariant $\tilde{z}$ shows no clustering by lighting but groups by subject identity.}
\label{fig:tsne_eyb}
\end{figure*}

\subsection{Practicalities of $\varepsilon$-noise and bottleneck-mask}
\label{subsec:epsilon_practical}

With exact computation, information is only lost when $m_i = 0$ because scaling by $m_i \in (0,1)$ is an isomorphic map. In real computation, however, multiplication operations are not isomorphic due to imprecision in floating point arithmetic. These imprecisions, alongside commonly undertaken computational procedures (e.g., clipping) induce a non-trivial forgetting region, under which $z_i$ with ``reasonable'' variance may be forgotten. Thus, practically speaking, we \emph{do not need to add $\varepsilon$-noise artificially} to erase information even when $m_i \neq 0$. Hence, we do not need $m_i$ to be exactly zero to lose information. The background noise of computation is sufficient for $I(\tilde{z} : z)$ to be smoothly controlled outside of zero $m_i$. The proposed framework implements exactly this control on $I(\tilde{z} : z)$, and thereby $I(\tilde{z} : x)$. Further, since it also optimizes a distortion measure between $\tilde{z}$ and $y$, this forms an information bottleneck~\cite{bib:vib}. For categorical $y$ with cross-entropy loss, minimization of $m_i$ coupled with the prediction task is equivalent to the bottleneck objective from~\cite{bib:ib}.

The goal of this work, however, is to generate invariant representations and not necessarily optimal bottleneck embeddings. In order to induce invariance to specific $s$, we learn the parameters of the bottleneck (forget-gate) so that it filters this information out of $z$ (mechanism described in Section~\ref{sec:method}). This encourages the encoder and the forget-gate to generate $\tilde{z}$ with minimal $I(\tilde{z} : s)$. The adversary operates only on $m$ and thus can be thought of as optimizing element-wise the channel between $z$ and $\tilde{z}$, minimizing $I(\tilde{z} : s)$ (or the equivalent general co-dependence term).

\begin{table}
\centering
\caption{\label{tab:eyb}Extended Yale-B results (random chance $s = 0.2$)}
\begin{tabular}{lcc}
\toprule
\textbf{Model} & $A_y$ & $A_s$ \\
\cmidrule(r){1-1} \cmidrule(lr){2-2} \cmidrule(l){3-3}
NN+MMD & 0.82 & -- \\
VFAE & 0.85 & 0.57 \\
CAI & 0.89 & 0.57 \\
CVIB & 0.82 $\pm$ 0.01 & 0.45 $\pm$ 0.03 \\
UAI & \textbf{0.95} & 0.24 \\
\cmidrule(r){1-1}
\textbf{Ours} & \textbf{0.95 $\pm$ 0.01} & \textbf{0.20 $\pm$ 0.01} \\
$\Delta$ over UAI & -- & 100\% \\
\bottomrule
\end{tabular}
\end{table}

The masks generated by the forget-gate have an intuitive interpretation: for each component, it either allows information to pass from $z_i$ to $\tilde{z}_i$ or does not. The overall design of the proposed framework causes separation of factors of $x$ that are correlated with $s$ from those that are not, so that they occur in different components of $z$, allowing a component-wise mask to effectively include or exclude factors from the final representation $\tilde{z}$ without cross-factor considerations. Removal of nuisance $s$ does not penalize the training objective (Equation~\ref{eq:objective}). However, for biasing $s$ correlated with $y$, the forget-gate will choose whether to allow their inclusion in $\tilde{z}$ based on the loss-weights in the objective.

\section{Experimental Evaluation}
\label{sec:evaluation}

The proposed framework is compared with NN+MMD, VFAE, CAI, CVIB and UAI. Performance is evaluated on two metrics: accuracy of predicting $y$ from $\tilde{z}$ ($A_y$) using the jointly trained predictor and that of predicting $s$ from $\tilde{z}$ ($A_s$) using a two-layer neural network trained \textit{post hoc}. While a high $A_y$ is desired, for true invariance $A_s$ should be random chance for nuisances and the share of the majority $s$-class for biasing $s$. Mean and standard deviation are reported based on five runs, except when results are quoted from previous works. Relative improvements in error-rate ($\Delta$) are also reported with error-rate for $A_s$ defined as the gap between the observed $A_s$ and its optimal value. We further report evaluation results of the framework in a multi-task setting.

\begin{figure*}
\centering
\includegraphics[width=\tsnefs\textwidth]{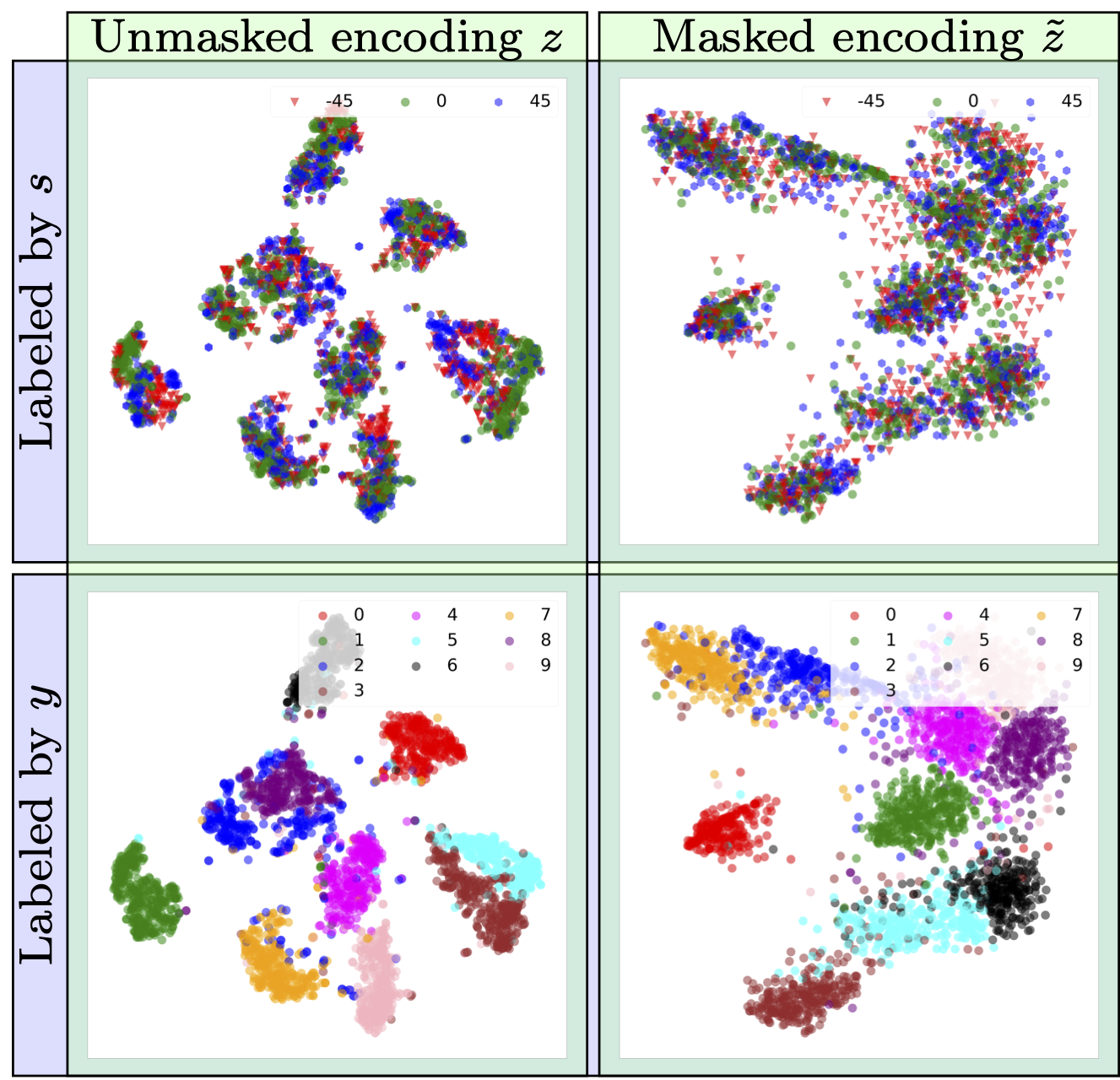}
\caption{MNIST-ROT -- t-SNE visualization of $z$ and $\tilde{z}$ labeled with rotation-angle ($s$) and digit-class ($y$). The invariant $\tilde{z}$ shows no clustering by $s$ while $z$ shows clear $s$-subgroups within each $y$-cluster.}
\label{fig:tsne_mnist}
\end{figure*}

\subsection{Robustness through invariance to nuisances}

Invariance to nuisances is evaluated on the Chairs~\cite{bib:uai}, Extended Yale-B~\cite{bib:eyb}, and MNIST-ROT~\cite{bib:uai} datasets. The network architectures for the forget-gate and the encoder are kept the same in all experiments. Besides quantitative results, we show t-SNE plots of $z$ and $\tilde{z}$ to visualize the transformation of the latent space due to $\tilde{z} = z \odot m$. These plots show that invariance is indeed brought about by the adversarial forgetting operation.

\sssection{Chairs.} This is a dataset of 86,366 images of 1,393 types of chairs at $31$ yaw and two pitch angles. We use the same version of this dataset as prior works, which is split into training and testing sets by picking alternate yaw angles, such that there is no overlap of angles between the two sets. The chair type is treated as $y$ and the yaw binned into four classes (front, left, right and back) as $s$. We use the same architecture for the encoder, the predictor, and the decoder as UAI~\cite{bib:uai}, i.e., two-layer networks. The discriminator is modeled as a two-layer network. Results of our experiments are summarized in Table~\ref{tab:chairs}. The proposed model achieves large improvements over the prior state-of-the-art (UAI) on both $A_y$ and $A_s$, with $A_s$ that is exactly random chance. Figure~\ref{fig:tsne_chairs} shows the t-SNE visualization of $z$ and $\tilde{z}$, exhibiting that $z$ groups by orientation but $\tilde{z}$ does not.

\sssection{Extended Yale-B.} This is a dataset of face-images of $38$ subjects captured under various lighting conditions. The prediction target $y$ is the subject-ID, while the nuisance $s$ is the lighting condition binned into five classes (four corners and frontal). For each subject, one image from each $s$-category is used for training and the rest of the dataset is used for testing~\cite{bib:uai,bib:vfae,bib:cai}. We use the same architecture for the encoder and the predictor as previous works~\cite{bib:uai,bib:cai}, i.e., one layer for each of these modules. The decoder and the discriminator are modeled with two layers each. Results of our experiments are shown in Table~\ref{tab:eyb}. The proposed model exhibits state-of-the-art performance on both $A_y$ and $A_s$, with $A_s$ being exactly random chance. This shows that the proposed model is able to completely remove information about the lighting direction from the latent embedding while retaining high $A_y$. Figure~\ref{fig:tsne_eyb} shows the t-SNE visualizations of $z$ and $\tilde{z}$ and validates this claim as the clustering of the latent embedding completely changes from grouping by $s$ for $z$ to grouping by $y$ for $\tilde{z}$.

\begin{table*}
\centering
\caption{\label{tab:mnist}MNIST-ROT results (random chance of $s = 0.2$). $\Theta$ represents the angles seen during training, i.e., $\{0, \pm 22.5^{\degree}, \pm 45^{\degree}\}$.}
\begin{tabular}{lcccc}
\toprule
\textbf{Model} & \multicolumn{3}{c}{$A_y$} & $A_s$ \\
\cmidrule(lr){2-4} \cmidrule(l){5-5}
& $\Theta$ & $\pm 55^{\circ}$ & $\pm 65^{\circ}$ & $\Theta$ \\
\cmidrule(r){1-1} \cmidrule(lr){2-4} \cmidrule(l){5-5}
NN+MMD & 0.970 $\pm$ 0.001 & 0.831 $\pm$ 0.001 & 0.665 $\pm$ 0.002 & 0.380 $\pm$ 0.011 \\
VFAE & 0.953 $\pm$ 0.004 & $\times$ & $\times$ & 0.389 $\pm$ 0.076 \\
CAI & 0.958 & 0.829 & 0.663 & 0.384 \\
CVIB & 0.960 $\pm$ 0.008 & 0.819 $\pm$ 0.007 & 0.674 $\pm$ 0.009 & 0.382 $\pm$ 0.005 \\
UAI & 0.977 & 0.856 & 0.696 & 0.338 \\
\cmidrule(r){1-1}
\textbf{Ours} & \textbf{0.991 $\pm$ 0.001} & \textbf{0.863 $\pm$ 0.001} & \textbf{0.730 $\pm$ 0.001} & \textbf{0.201 $\pm$ 0.001} \\
$\Delta$ over UAI & 60.9\% & 4.86\% & 11.18\% & 99.3\% \\
\bottomrule
\end{tabular}
\end{table*}

\sssection{MNIST-ROT.} This is a variant of the MNIST~\cite{bib:mnist} dataset, which is augmented with digits rotated at $\pm 22.5^{\degree}$ and $\pm 45^{\degree}$. The digit class is treated as $y$ and the rotation angle as categorical $s$. We use the same architecture for the encoder, the predictor, and the decoder as UAI, i.e., two layers for the encoder and the predictor each and three layers for the decoder. The discriminator is modeled with two layers. Table~\ref{tab:mnist} summarizes the results of our experiments on test images with rotation angles both seen and unseen ($\pm 55^{\degree}$ and $\pm 65^{\degree}$) during training. Our model achieves not only the best $A_y$ but also $A_s$ that is exactly random chance, showing that it is able to filter out $s$ while retaining more information about $y$, leading to more accurate $y$-predictions. The t-SNE visualizations of $z$ and $\tilde{z}$ shown in Figure~\ref{fig:tsne_mnist} further validate this as $\tilde{z}$ is clustered by $y$ but has uniformly distributed $s$, while $z$ shows distinct groups of $s$ in each digit cluster.

\subsection{Fairness through invariance to biasing factors}

The Adult~\cite{bib:ag} and German~\cite{bib:ag} datasets have been popularly employed in fairness settings~\cite{bib:nnmmd,bib:vfae,bib:cvib,bib:cai}. The former is an income dataset for classifying whether a person has more than \$50,000 savings and the biasing $s$ is age. The latter is used for predicting whether a person has a good credit rating, and gender is the biasing $s$. We use architectures similar to VFAE for encoders (two layers), predictors (one layer), and decoders (two layers), along with a two-layer discriminator. Results of experiments on these datasets are shown in Tables~\ref{tab:adult} and \ref{tab:german}. The proposed model completely removes information about $s$, as reflected by $A_s$ being the same as the population share of the majority $s$-class for both the datasets. Previous works have also achieved perfect score on $A_s$ but the proposed model outperforms them on $A_y$ on both the datasets, showing that it is able to retain more information for predicting $y$ while being fair by successfully filtering only biasing factors.

\begin{table}
\centering
\caption{\label{tab:adult}Adult results (majority class of $s = 0.67$).}
\begin{tabular}{lcc}
\toprule
\textbf{Model} & $A_y$ & $A_s$ \\
\cmidrule(r){1-1} \cmidrule(lr){2-2} \cmidrule(l){3-3}
NN+MMD & 0.75 $\pm$ 0.00 & \textbf{0.67 $\pm$ 0.01} \\
VFAE & 0.76 $\pm$ 0.01 & \textbf{0.67 $\pm$ 0.01} \\
CAI & 0.83 & 0.89 \\
CVIB & 0.69 $\pm$ 0.01 & 0.68 $\pm$ 0.01 \\
\cmidrule(r){1-1}
\textbf{Ours} & \textbf{0.85 $\pm$ 0.00} & \textbf{0.67 $\pm$ 0.00} \\
$\Delta$ over VFAE & 37.5\% & -- \\
\bottomrule
\end{tabular}
\end{table}

\subsection{Invariance in multi-task learning}

The proposed framework is evaluated on the dSprites~\cite{bib:dSprites} dataset of shapes with independent factors: color, shape, scale, orientation, and position. The dataset was preprocessed following~\cite{bib:scan}, resulting in two classes for scale and four for position and orientation each. Shape ($y^{(1)}$) and scale ($y^{(2)}$) are treated as the prediction tasks, where shape is desired to be invariant to position ($s^{(1)}$) and scale to orientation ($s^{(2)}$). We use the same component networks that we used for Extended Yale-B. We compare results with a version of our model without the decoder, the maskers and the discriminators, i.e., both $y^{(1)}$ and $y^{(2)}$ are predicted from $z$. Evaluation could not be conducted on NN+MMD, VFAE, CAI and CVIB because they have one $\tilde{z}$ and all tasks have to be invariant to the same $s$, and on UAI because it works only for a single $y$. Table~\ref{tab:dSprites} presents the results. The proposed framework achieves the same accuracies for $y^{(1)}$ and $y^{(2)}$ as the baseline, while maintaining random chance accuracies for $s^{(1)}$ ($0.5$) and $s^{(2)}$ ($0.25$) as opposed to significantly higher corresponding scores for the baseline. Hence, our framework works effectively in multi-task settings.

\section{Conclusion}
\label{sec:conclusion}

We have presented a novel framework for invariance induction in neural networks through ``forgetting'' of information related to unwanted factors. We showed that the forget-gate used in the proposed framework acts as an information bottleneck and that adversarial training encourages generation of forget-masks that remove unwanted factors. Empirical results show that the proposed model exhibits state-of-the-art performance in both nuisance and bias settings.

\begin{table}
\centering
\caption{\label{tab:german}German results (majority class of $s = 0.8$).}
\begin{tabular}{lcc}
\toprule
\textbf{Model} & $A_y$ & $A_s$ \\
\cmidrule(r){1-1} \cmidrule(lr){2-2} \cmidrule(l){3-3}
NN+MMD & 0.74 $\pm$ 0.01 & \textbf{0.80 $\pm$ 0.00} \\
VFAE & 0.70 $\pm$ 0.00 & \textbf{0.80 $\pm$ 0.00} \\
CAI& 0.70 & 0.81 \\
CVIB & 0.74 $\pm$ 0.00 & \textbf{0.80 $\pm$ 0.00} \\
\cmidrule(r){1-1}
\textbf{Ours} & \textbf{0.76 $\pm$ 0.00} & \textbf{0.80 $\pm$ 0.00} \\
$\Delta$ over CVIB & 7.7\% & -- \\
\bottomrule
\end{tabular}
\end{table}

\begin{table}
\centering
\caption{\label{tab:dSprites}Results on dSprites in multi-task setting. Random chance of $s$ for task \#1 is $0.5$ and for task \#2 is $0.25$.}
\begin{tabular}{lcccc}
\toprule
& \multicolumn{2}{c}{\textbf{Task \#1}} & \multicolumn{2}{c}{\textbf{Task \#2}} \\
\cmidrule(lr){2-3} \cmidrule(l){4-5}
\textbf{Acc} & \textbf{Baseline} & \textbf{Ours} & \textbf{Baseline} & \textbf{Ours} \\
\cmidrule(r){1-1}  \cmidrule(lr){2-3} \cmidrule(l){4-5}
$A_y$ & 0.99 & 0.99 & 0.99 & 0.99 \\
$A_s$ & 0.94 & \textbf{0.50} & 0.40 & \textbf{0.25} \\
\bottomrule
\end{tabular}
\end{table}

\vspace{5pt}
\noindent\textbf{Acknowledgements.} This work is based on research sponsored by the Defense Advanced Research Projects Agency (agreement number FA8750-16-2-0204). The U.S. Government is authorized to reproduce and distribute reprints for governmental purposes  notwithstanding any copyright notation thereon. The views and conclusions contained herein are those of the authors and should not be interpreted as necessarily representing the official policies or endorsements, either expressed or implied, of the Defense Advanced Research Projects Agency or the U.S. Government.

\appendix
\section{Variance Inequality}
\label{sec:app:var-id}

For two random variables $A$ and $B$, possibly dependent,
{
\mathfsapp
\begin{flalign}
    &\text{Cov}(A,B) \leq \sqrt{\text{Var}(A)\text{Var}(B)} \leq \frac{1}{2} (\text{Var}(A) + \text{Var}(B))))&&&&& \label{app:eq:geo-arth} \\
    &\text{Var}(A + B) = \text{Var}(A) + \text{Var}(B) + 2\text{Cov}(A,B) \leq 2 \big (\text{Var}(A) + \text{Var}(B) \big )&&&&&
    \raisetag{1\normalbaselineskip}
\end{flalign}
}
Equation~\ref{app:eq:geo-arth} holds due to the geometric mean--arithmetic mean inequality. Let $A=(m_i - \mathbb{E}[m_i]) z_i$ and $B=\mathbb{E}[m_i]z_i$. Then, $\text{Var}(m_i z_i) = \text{Var}(A + B)$ gives:
{
\mathfsapp
\begin{align}
    \text{Var}(m_i z_i) \leq 2\text{Var}((m_i - \mathbb{E}[m_i]) z_i) + 2 \mathbb{E}[m_i]^2\text{Var}(z_i)
\end{align}
}

{
\bibliographystyle{aaai}
\fontsize{9.5pt}{10.5pt}\selectfont
\bibliography{main}
}

\end{document}